\documentclass[runningheads]{llncs}
\usepackage{xcolor}
\usepackage[T1]{fontenc}
\usepackage{graphicx}
\usepackage{booktabs}
\usepackage{amsmath}
\usepackage{amssymb}
\usepackage{microtype}
\usepackage[hyphens]{url}
\usepackage{hyperref}
\usepackage{tikz}
\usepackage{pgfplots}
\pgfplotsset{compat=1.18}
\usepgfplotslibrary{groupplots}
\usetikzlibrary{calc}
\hypersetup{
  colorlinks=true,
  linkcolor=blue!50!black,
  citecolor=blue!50!black,
  urlcolor=blue!50!black,
}
\definecolor{cblue}{HTML}{1F77B4}
\definecolor{cred}{HTML}{D62728}
\definecolor{corange}{HTML}{FF7F0E}
\definecolor{cgreen}{HTML}{2CA02C}
\definecolor{cpurple}{HTML}{9467BD}
\definecolor{cgray}{HTML}{7F7F7F}
\definecolor{cgraylight}{HTML}{BDBDBD}

\begin{document}

\title{Before and After Temperature: A Distributional View
       of Creative LLM Generation}
\titlerunning{Before and After Temperature}

\author{V.S. Raghu Parupudi\inst{1,2} \and
Harsha Ponnada\inst{2} \and
Aditi Kaushal\inst{3} \and
S. Shria Parupudi\inst{4} \and
Saiteja Dasari\inst{2,5} \and
Sahiti Bulusu\inst{6}}
\authorrunning{V.S. Raghu Parupudi et al.}
\institute{University of California, San Diego \and
Indian Institute of Technology, Kharagpur \and
Delhi Technological University \and
Carnegie Mellon University \and
Rutgers University \and
Georgia Institute of Technology}

\maketitle

\begin{abstract}

Reference-free evaluation of large language model (LLM) creativity
relies on  perplexity, entropy, and top-1 margin. We show that a much stronger signal lives one
step earlier in the pipeline: in how sampling temperature
\emph{reshapes} the model's token distribution before the next token
is drawn. On Llama-3.1-8B-Instruct generations of 500 open-ended
creative prompts at $T \in \{0.3, 0.8, 1.5\}$, a single per-token
feature derived from this reshaping predicts the within-prompt
creativity rank at Spearman $\rho{=}0.918$ against an averaged
gpt-4o\,/\,gemini-2.5-pro judge ($n{=}500$) and $\rho{=}0.870$
against a three-rater human-majority ranking ($n{=}150$). Each of
four standard reference-free baselines (self-perplexity, mean
predictive entropy, top-1 margin, gzip compression ratio) tops out
at $|\rho|\!\approx\!0.76$ on both ground truths: a gap of $+0.165$
on averaged-LLM and $+0.110$ on human-majority, both far larger
than the spread among the baselines themselves.
The two ground-truth panels agree with each other at $\rho{=}0.83$,
above the inter-human ceiling of $\rho{=}0.77$, so the comparison is
not bottlenecked by judge noise. Mechanistically, the win comes from
a sharp distributional signature of the incoherence regime: at
$T{=}1.5$ the cumulative-mass width $n_{95}(q)$ inflates from
$\sim\!1$ to ${\sim}\!131$ tokens and post-temperature mass leaks
off the pre-temperature top-$90\%$ plausible set by about
$13$ percentage points. The
per-token aggregates do not separate $T{=}0.8$ from $T{=}0.3$;
discriminating the two coherent regimes is left to sequence-level
features.

\keywords{
LLM evaluation \and reference-free metrics \and LLM-as-judge
\and sampling temperature \and distributional analysis.}
\end{abstract}

\section{Introduction}
\label{sec:intro}

How do we evaluate a language model's output when there's no
reference text to compare against, as in open-ended creative
generation? Perplexity is still the default \cite{brown1993statistical,jelinek1977perplexity}, but a generation of work has shown it falls
short on open-ended tasks. Low perplexity, on its own, rewards fluent
and high-frequency text. That is exactly the failure mode Holtzman
et~al.\ \cite{holtzman2020curious} called ``neural text degeneration''.
Reference-free quality metrics built on model embeddings or
LLM-as-judge prompts have closed some of these gaps
\cite{zhang2020bertscore,liu2023geval,fu2024gptscore}, but they work
on the \emph{output text}, not on the \emph{distribution} the model
produced it from.

A separate line of work treats the model's own distribution as a
signal: typicality \cite{meister2023typical}, semantic entropy for
hallucination detection \cite{kuhn2023semantic,farquhar2024detecting}, self-consistency
\cite{wang2023selfconsistency,manakul2023selfcheckgpt}, and learned
``confidence'' scores \cite{parupudi2025confidence}. These either
collapse the distribution to a scalar (entropy, perplexity, margin)
or compare across multiple samples. Neither approach uses the fact
that sampling temperature is itself a \emph{transformation} applied
to the model's raw distribution before any token is emitted. At every
generation step there are two distributions: the model's true belief
$p \propto \exp(z)$ and the temperature-reshaped distribution $q
\propto \exp(z/T)$ from which the next token is actually sampled.
The gap between $p$ and $q$ is computable from the same forward pass
that produces the text.

A natural question is to ask, \textbf{does the structure of this
pre-vs-post temperature gap tell us how good the generation is?}

We answer empirically. We generate at $T \in \{0.3, 0.8, 1.5\}$ on
Llama-3.1-8B-Instruct \cite{grattafiori2024llama3} for 500 creative
prompts drawn from published sources (WritingPrompts
\cite{fan2018hierarchical}, the Alternative Uses Task
\cite{guilford1967nature,stevenson2022putting}, and HellaSwag
\cite{zellers2019hellaswag}), capturing the raw top-$K$ logits at
every generation step on a GCP L4 GPU ($K{=}100$ for
$T\!\leq\!0.8$, $K{=}200$ for $T{=}1.5$, with a guarantee that the
sampled token is always in the captured set).

Our findings, ordered by what we think matters most:
(i)~a single per-token feature derived from the pre-vs-post
temperature reshape correlates with the within-prompt creativity
rank at $\rho{=}0.918$ against an averaged LLM judge and
$\rho{=}0.870$ against the human majority, beating each of four
standard reference-free baselines by more than $0.10$ in absolute
$\rho$ on both ground truths; (ii)~the underlying mechanism is a
sharp distributional signature of the incoherence regime
($n_{95}(q)$ inflation, ${\sim}13$-pp mass leakage off the
pre-temperature top-$90\%$ plausible set); and (iii)~per-token aggregates do not separate
$T{=}0.8$ from $T{=}0.3$, which we read as a pointer toward
sequence-level features for that finer distinction rather than as a
limitation of the pre-vs-post view itself.

\section{Related Work}
\label{sec:related}

\paragraph{Reference-free quality metrics.}
Perplexity \cite{brown1993statistical,jelinek1977perplexity} is still
widely used as a quality stand-in, even though its problems are well
known. Holtzman et~al.\ \cite{holtzman2020curious} showed that
minimising perplexity drives ``neural text degeneration'': bland,
repetitive output that hits a fluency target while saying nothing.
Fabbri et~al.\ \cite{fabbri2021summeval} found that perplexity-class
metrics correlate weakly with human quality judgements on
summarisation. Reference-based n-gram metrics like BLEU
\cite{papineni2002bleu} simply do not apply when there is no
reference. Embedding-based reference-free signals such as BARTScore
\cite{yuan2021bartscore} and BERTScore variants
\cite{zhang2020bertscore} have filled that gap for open-ended tasks,
but they score the surface form of the output, not the distribution
that produced it. Our work targets that second layer directly.

\paragraph{LLM-as-judge.}
A second strand of work treats evaluation as a generation task in its
own right: prompt a strong LLM to score or rank candidates. G-Eval
\cite{liu2023geval} and GPTScore \cite{fu2024gptscore} are the
canonical examples; Prometheus \cite{kim2024prometheus} is an
open-weight judge in the same family. These judges correlate well
with humans on many tasks but can over-reward fluent incoherence
when an answer opens strongly and falls apart later; Chakrabarty
et~al.\ \cite{chakrabarty2024art} report this failure mode on
long-form creative writing. We use a coherence-first ranking
prompt to control for it (Section~\ref{sec:ground-truth}).

\paragraph{Model-internal uncertainty.}
A third line of work mines the generating model's own distribution
for signal. Predictive entropy and margin-based confidence are the
classical handles \cite{malinin2021uncertainty}. Semantic entropy
\cite{kuhn2023semantic} clusters multiple sampled completions by
meaning, and the same idea drives hallucination detection
\cite{farquhar2024detecting}. Locally typical decoding
\cite{meister2023typical} is the closest in spirit to what we do:
it treats each token's deviation from expected surprisal as a signal,
but uses it to pick tokens during sampling rather than to evaluate
afterwards. Self-consistency \cite{wang2023selfconsistency} and
SelfCheckGPT \cite{manakul2023selfcheckgpt} use agreement across
multiple samples, which works but multiplies inference cost. All of
these methods collapse the distribution to a scalar or compare across
samples. None of them looks at the structure of how temperature
itself reshapes the distribution at a single step.

\paragraph{Diversity and degeneration.}
Corpus-level diversity has been measured with distinct-$n$
\cite{li2016diversity}, self-BLEU \cite{zhu2018texygen}, and
compression-ratio variants \cite{shaib2024standardizing,padmakumar2024writing}. These tools tell you how varied a sample set
is in aggregate. They cannot tell you anything about the per-token
shape of the distribution behind any single response, which is the
level we operate at. We include compression ratio as one of our
sequence-level baselines (Section~\ref{sec:vs-baselines}) precisely
because it tests whether surface diversity alone explains what our
distributional features pick up. It does not.

\paragraph{Temperature as a study variable.}
Prior decoding work (nucleus \cite{holtzman2020curious}, typical
\cite{meister2023typical}) notes that the tail behaves differently
at high $T$, but does not quantify the pre-vs-post reshape itself.
That is the question we set out to answer.

\section{Experimental Setup}
\label{sec:setup}

\subsection{Prompts}
We select 500 open-ended creative prompts (English, 15--250
characters, no single-answer items) from three published sources:
\textbf{WritingPrompts} \cite{fan2018hierarchical} (168 story
starters from r/WritingPrompts, filtered for self-contained prompts);
the \textbf{Alternative Uses Task} \cite{guilford1967nature,stevenson2022putting} (166 ``list creative uses for $X$'' prompts,
the standard divergent-thinking instrument); and the
\textbf{HellaSwag} ActivityNet split \cite{zellers2019hellaswag} (166
scene-continuation prompts of the form ``continue this scene in a
vivid, imaginative way''). All sources are licensed for research
use.

\subsection{Model and Generation}
We generate all samples with \textbf{meta-llama/Llama-3.1-8B-Instruct}
\cite{grattafiori2024llama3} in fp16, on a single NVIDIA L4 GPU
(24~GB VRAM) on Google Cloud Platform.

For each prompt we generate three independent autoregressive traces,
one at each $T \in \{0.3, 0.8, 1.5\}$, with
\texttt{max\_new\_tokens}${=}300$ and the model's default chat
template. Contexts diverge from step 1 onward; we do not share
traces across temperatures.

\subsection{Logit Capture}
\label{sec:logit-capture}
For each generated token, we record the top-$K$ raw logits (token id
and logit value) before any sampling. The capture depth is
a function of temperature:

\begin{itemize}
  \item $T \in \{0.3, 0.8\}$: $K{=}100$. The post-temperature
        distribution is highly peaked, and the top-100 captures
        $\geq 0.99$ of the mass for virtually all positions.
  \item $T = 1.5$: $K{=}200$. The post-temperature distribution is
        much flatter; we need a wider head to retain
        $\geq 0.95$ of the mass for stable divergence computation.
\end{itemize}

\begin{table}[h]
\centering
\caption{Temperature for Top-$K$ logits capture sufficiency. The sampled-token
append guarantee (Section~\ref{sec:logit-capture}) is load-bearing at $T{=}1.5$:
in $88.4\%$ of positions the actually sampled token lay outside the captured
top-$K{=}200$ and had to be appended. This confirms that $K{=}200$ at $T{=}1.5$ is
justified but tight.}
\label{tab:logit-capture}
\begin{tabular}{lcrrr}
\toprule
$T$  & $K$ & Positions   & Appends   & Append rate \\
\midrule
0.3  & 100 & 144{,}774   & 0         & 0.00\%      \\
0.8  & 100 & 143{,}841   & 129       & 0.09\%      \\
1.5  & 200 & 149{,}222   & 131{,}927 & 88.41\%     \\
\midrule
Total & --- & 437{,}837  & 132{,}056 & ---         \\
\bottomrule
\end{tabular}
\end{table}

We additionally guarantee that the \emph{sampled} token is always
in the recorded list (appended if it lay outside top-$K$), so that
chosen-token analyses remain well-defined at high $T$. The append
rate (Table~\ref{tab:logit-capture}) jumps roughly $1000\times$ from
$T{=}0.8$ to $T{=}1.5$, confirming that $K{=}200$ at $T{=}1.5$ is
reasonable but tight and that the append is load-bearing.

\subsection{Two Distributions per Token}
\label{sec:two-dists}
We now have access to the raw
logits $z$ (top-$K$) at each generated token of each trace. From these we calculate on the renormalised
top-$K$ head:

\begin{itemize}
  \item \emph{Pre-temperature distribution}: $p =
        \mathrm{softmax}(z)$, the model's untouched belief about
        next-token candidates at this position.
  \item \emph{Post-temperature distribution}: $q =
        \mathrm{softmax}(z / T)$, the temperature-reshaped
        distribution from which the actual sampled token was drawn.
\end{itemize}

At $T{=}1$ exactly, $p{\equiv}q$ and every divergence vanishes; we
exclude it from the main grid for that reason, and exclude $T{=}0$
because greedy decoding degenerated visibly. The construction is
identical across all three temperatures, so within-trace and
within-prompt comparisons are exact. We use divergences only to rank
prompts and traces relative to each other, not against any external
benchmark.

\subsection{Ground Truth}
\label{sec:ground-truth}
We use two independent layers of ground truth.

\paragraph{LLM-judge ensemble (full 500 prompts).}
For every prompt, we present its three responses to two judges from
different model families: \textbf{gpt-4o} \cite{achiam2023gpt} (via
the OpenAI API) and \textbf{gemini-2.5-pro} (via Google Vertex AI).
The judge prompt is identical across the two models. It asks for a
three-way ranking by creativity with ties allowed, explicitly
penalizes incoherence, code-switching, and disconnected technical
jargon, and instructs the judge to read the whole response rather
than just the opening. The presentation order of the three responses
is randomly shuffled per prompt (seeded), so position bias is
averaged out across the corpus. Each judge produces 500 rankings.

\paragraph{Human raters (150-prompt subset).}
The first 150 prompts (56 WP, 48 HellaSwag, 46 AUT) were ranked by
three independent raters, co-authors of this paper, working
blind to temperature labels and without communicating during
annotation, under the same coherence-first rule as the LLM
judges and a fixed A/B/C presentation order so judgements are
directly comparable. Inter-rater agreement is reported in
Section~\ref{sec:gt-rankings}.
\section{Distributional Metrics}
\label{sec:metrics}

Given the per-token pre- and post-temperature distributions $p$ and $q$
from Section~\ref{sec:two-dists}, we compute a set of measures, each
capturing a different aspect of how temperature reshapes the
distribution. They fall into five families.

\subsection{Information-Theoretic Divergences}

\begin{itemize}
  \item $\mathrm{KL}(q \| p) = \sum_i q_i \log (q_i / p_i)$:
        post-to-pre Kullback--Leibler divergence.
  \item $\mathrm{KL}(p \| q)$: the reverse direction. At $T{=}0$ this
        blows up to $+\infty$ because $q$ is one-hot. We do not use
        $T{=}0$ in the main grid, but we keep the metric to expose the
        asymmetry.
  \item $\mathrm{JS}(p, q) = \tfrac{1}{2} \mathrm{KL}(p \| m) +
        \tfrac{1}{2} \mathrm{KL}(q \| m)$ with $m = (p+q)/2$:
        Jensen--Shannon, symmetric and bounded.
\end{itemize}

\subsection{Geometric Distances}
\begin{itemize}
  \item Total variation: $\mathrm{TV}(p, q) = \tfrac{1}{2}
        \sum_i |p_i - q_i|$.
  \item $L_2$: $\sqrt{\sum_i (p_i - q_i)^2}$.
  \item Hellinger: $\tfrac{1}{\sqrt{2}}
        \sqrt{\sum_i (\sqrt{p_i} - \sqrt{q_i})^2}$.
\end{itemize}

\subsection{Top-end Properties}
\begin{itemize}
  \item Top-1 probability under pre and post: $\max_i p_i$, $\max_i
        q_i$.
  \item Top-1 shift: $\max_i q_i - \max_i p_i$. Positive at
        $T \!<\! 1$ (sharpening promotes the leader), negative at
        $T \!>\! 1$ (flattening demotes it).
  \item Top-1 identity change: indicator that the $\arg\max$
        differs between $p$ and $q$.
  \item Margin: $p_{(1)} - p_{(2)}$ (and the same for $q$).
\end{itemize}

\subsection{Chosen-Token Properties}
At each step we record the token that was actually sampled. By
construction it sits in the captured top-$K$ (Section~\ref{sec:logit-capture}).
\begin{itemize}
  \item $p[\text{chosen}]$: the pre-temperature probability of the
        sampled token, i.e.\ how plausible was this pick on the raw
        distribution? High values mean the chosen token was already
        a leader; low values mean temperature promoted a long-tail
        candidate.
  \item $q[\text{chosen}]$: the post-temperature probability of the
        sampled token.
\end{itemize}

\subsection{Cumulative-Mass Width and Mass Leakage}
Two complementary measures of how temperature spreads mass across
candidates.

\paragraph{Top-$n_{\alpha}$ width.} For threshold
$\alpha \in \{0.90, 0.95, 0.99\}$, $n_\alpha(p)$ is the smallest number
of top-ranked tokens whose probability sums to at least $\alpha$. We
compute $n_\alpha(p)$ and $n_\alpha(q)$ at every step. The signed
difference $n_\alpha(q) - n_\alpha(p)$ is the \emph{flattening}: positive
when temperature spreads mass into more candidates, negative when it
concentrates.

When the captured top-$K$ head sums to less than $\alpha$, which only
happens at $T{=}1.5$ for $\alpha{=}0.99$ on a small fraction of
positions, we mark $n_\alpha$ as censored at $K$ and drop those positions
from analyses that need an uncensored value.

\paragraph{Mass leakage.} Let $S_\alpha^{(p)}$ be the set of tokens that
make up $p$'s top-$n_\alpha(p)$ mass, i.e.\ the ``pre-temperature
$\alpha$-plausible set''. Then:
\begin{equation}
\textsc{PostCumOnPre}_{\alpha} \;=\; \sum_{i \in S_\alpha^{(p)}}
q_i.
\end{equation}
This is the post-temperature mass that stays on the original plausible
set. Anchor values:
\begin{itemize}
  \item At $T{=}1$ it equals $\alpha$ by construction.
  \item At $T{<}1$ it rises toward $1$: sharpening pushes more mass
        onto the original leaders.
  \item At $T{>}1$ it falls below $\alpha$: flattening leaks mass
        \emph{off} the originally-plausible set and onto the tail.
\end{itemize}
The symmetric quantity $\textsc{PreCumOnPost}_{\alpha} = \sum_{i \in
S_\alpha^{(q)}} p_i$ tells us the other direction: how much
pre-temperature mass was already on the set that ended up plausible
under temperature. Together they describe the redistribution.

\subsection{Per-Response Aggregation}
Each trace has about $290$ rows in the per-token table. For each metric
we aggregate per (prompt, $T$) using \texttt{mean, median, p10, p90,
max}, plus \texttt{min} for the chosen-token measures (which catches the
least-plausible commitment). We use multiple aggregators on purpose:
the same divergence looks different at the typical-token vs.\
worst-token level, and Section~\ref{sec:incoherence-signature} shows
that the $\mathrm{p10}$ aggregator (calmest position in the trace) is
often the strongest single signal.

\subsection{Audit of Degenerate Metrics}
Top-1 identity change is always $0$ within a step (rank-invariance);
$n_\alpha$ censoring is $0\%$ at $\alpha \in \{0.90, 0.95\}$ for our
chosen $K$; $\mathrm{KL}(p \| q)$ at $T{=}0$ is undefined. The
analysis in Section~\ref{sec:results} excludes these; the full
pre-audit catalogue is available on request.
\section{Results}
\label{sec:results}

\subsection{Ground-Truth Rankings}
\label{sec:gt-rankings}

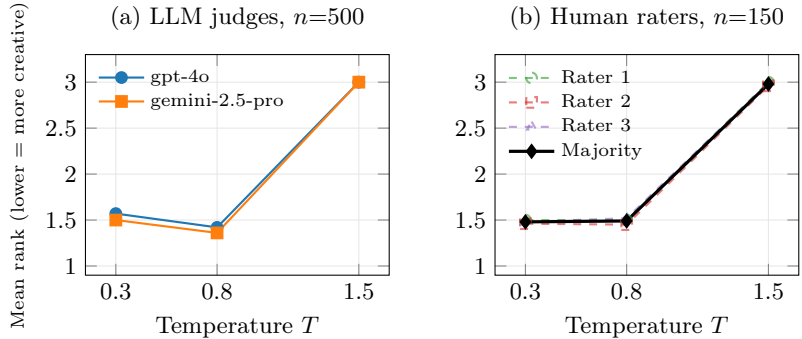
\begin{figure}[t]
\centering
\begin{tikzpicture}
\begin{groupplot}[
  group style={group size=2 by 1, horizontal sep=1.4cm},
  width=5.6cm, height=4.5cm,
  xlabel={Temperature $T$}, xtick={0.3, 0.8, 1.5},
  xmin=0.15, xmax=1.65, ymin=0.9, ymax=3.3,
  ytick={1, 1.5, 2, 2.5, 3},
  grid=major, grid style={gray!20, line width=0.3pt},
  tick label style={font=\footnotesize},
  label style={font=\footnotesize},
  title style={font=\footnotesize},
  legend cell align=left,
  legend style={font=\scriptsize, draw=none, fill=none,
                at={(0.03,0.97)}, anchor=north west,
                inner sep=1pt, row sep=-1pt},
  every axis plot/.append style={thick, mark size=2pt},
]
\nextgroupplot[title={(a) LLM judges, $n{=}500$},
               ylabel={Mean rank (lower $=$ more creative)},
               ylabel style={font=\scriptsize}]
\addplot[color=cblue, mark=*]         coordinates {(0.3, 1.57) (0.8, 1.42) (1.5, 3.00)};
\addlegendentry{gpt-4o}
\addplot[color=corange, mark=square*] coordinates {(0.3, 1.50) (0.8, 1.36) (1.5, 3.00)};
\addlegendentry{gemini-2.5-pro}

\nextgroupplot[title={(b) Human raters, $n{=}150$}]
\addplot[color=cgreen,  mark=o,        dashed, opacity=0.6] coordinates {(0.3, 1.50) (0.8, 1.50) (1.5, 3.00)};
\addlegendentry{Rater 1}
\addplot[color=cred,    mark=square,   dashed, opacity=0.6] coordinates {(0.3, 1.46) (0.8, 1.45) (1.5, 2.96)};
\addlegendentry{Rater 2}
\addplot[color=cpurple, mark=triangle, dashed, opacity=0.6] coordinates {(0.3, 1.48) (0.8, 1.52) (1.5, 2.99)};
\addlegendentry{Rater 3}
\addplot[color=black,   mark=diamond*, very thick]          coordinates {(0.3, 1.48) (0.8, 1.49) (1.5, 2.98)};
\addlegendentry{Majority}
\end{groupplot}
\end{tikzpicture}
\caption{Mean rank by temperature for the two LLM judges (a) and
three human raters plus their majority (b). Lower is more creative
($1{=}$best of 3). Both panels show $T{=}1.5$ ranked last on
essentially every prompt; the mid-vs-bland call between $T{=}0.8$
and $T{=}0.3$ is close on both panels.}
\label{fig:ranking}
\end{figure}

\paragraph{LLM judge ensemble.}
Both gpt-4o and gemini-2.5-pro parsed cleanly on all 500 prompts.
Figure~\ref{fig:ranking}(a) shows the mean rank each judge assigned
to each temperature. gpt-4o ranked T${=}0.8$ first on 293 prompts
and T${=}0.3$ first on 217 (counts include ties at the top, so they
sum to more than 500). For gemini-2.5-pro the corresponding counts
are 321 and 248. Neither judge ranked T${=}1.5$ first on more than
one prompt out of 500. Both produce an inverted-U: T${=}0.8$ wins
on a majority of prompts, T${=}0.3$ is close behind, and T${=}1.5$
is ranked last on $499/500$ (gpt-4o) and $500/500$ (gemini-2.5-pro)
prompts.

Judge-judge agreement, measured as the mean Spearman rank
correlation across the three temperatures within each prompt, is
$\rho{=}+0.794$ (median $+0.866$). The judges are in lockstep at the
bottom of the ranking and disagree only on the bland-vs-mid ordering
at the top.

\paragraph{Human raters.}
On the 150-prompt subset, three independent raters ranked the
responses under the same coherence-first rule as the LLM judges.
Two raters finished all 150 items; Rater 3 finished 147.
Figure~\ref{fig:ranking}(b) shows mean rank per temperature for
each rater and for their mean-rank majority. All three raters
ranked T${=}1.5$ last on essentially every item (rater means
$\approx 3.00$). T${=}0.3$ and T${=}0.8$ are within $0.05$ rank
units of each other for every rater, which means the mid-vs-bland call is
the closer one even for humans.

Inter-rater agreement, measured per-prompt as Spearman across the
three temperatures, is $\rho{=}+0.798$ (Rater 1 vs Rater 3),
$\rho{=}+0.761$ (Rater 1 vs Rater 2), and $\rho{=}+0.755$ (Rater 2
vs Rater 3). The average is $\rho{=}+0.771$, which we take as the
human-agreement \emph{ceiling} for this task.

\subsection{Human-vs-LLM Agreement}
\label{sec:human-vs-llm}

\begin{figure}[t]
\centering
\begin{tikzpicture}[x=20cm, y=-0.5cm, font=\footnotesize]
\def\xmin{0.70}
\def\xmax{0.87}
\def\ceiling{0.771}

\foreach \xv in {0.70, 0.72, 0.74, 0.76, 0.78, 0.80, 0.82, 0.84, 0.86} {%
  \draw[gray!25, line width=0.3pt] (\xv, 0.3) -- (\xv, 6.5);%
}

\foreach \rho/\row/\colr/\lbl in {%
    0.761/0/cgray/Rater 1 vs Rater 2,%
    0.798/1/cgray/Rater 1 vs Rater 3,%
    0.755/2/cgray/Rater 2 vs Rater 3,%
    0.793/3/cblue/Maj.\ vs gpt-4o,%
    0.803/4/corange/Maj.\ vs gemini-2.5-pro,%
    0.832/5/black/Maj.\ vs avg(LLM)%
} {%
  \fill[fill=\colr] (\xmin, \row+0.7) rectangle (\rho, \row+1.3);%
  \node[anchor=east, font=\scriptsize] at (\xmin-0.002, \row+1) {\lbl};%
  \node[anchor=west, font=\scriptsize\bfseries]
    at (\rho+0.002, \row+1) {$+\pgfmathprintnumber[fixed,fixed zerofill,precision=3]{\rho}$};%
}

\draw[cgray, dotted, line width=1pt] (\ceiling, 0.3) -- (\ceiling, 6.5);
\node[cgray, font=\scriptsize\itshape, anchor=south, align=center]
  at (\ceiling, 0.3) {Inter-human ceiling\\$\bar\rho{=}0.771$};

\draw[gray!50, dashed, line width=0.4pt] (\xmin, 3.5) -- (\xmax+0.005, 3.5);

\draw[->, gray!70] (\xmin, 6.7) -- (\xmax+0.005, 6.7);
\foreach \xv in {0.70, 0.75, 0.80, 0.85} {%
  \draw[gray!70, line width=0.3pt] (\xv, 6.7) -- (\xv, 6.85);%
  \node[font=\scriptsize, anchor=north] at (\xv, 6.85) {\xv};%
}
\node[font=\footnotesize, anchor=north] at ({(\xmin+\xmax)/2}, 7.5)
  {Mean per-prompt Spearman $\rho$};
\end{tikzpicture}
\caption{Mean per-prompt Spearman $\rho$ for inter-human pairs
(gray, top) and human-majority vs each LLM judge (colored, bottom).
The averaged-LLM judge ($\rho{=}{+0.832}$) beats the inter-human
ceiling ($\bar\rho{=}{+0.771}$, dotted line) by a comfortable
margin.}
\label{fig:agreement}
\end{figure}
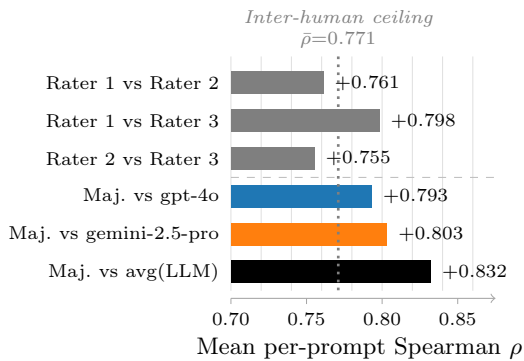

We compare each LLM judge, and their average, against the
human-majority ranking on the 150-prompt subset
(Figure~\ref{fig:agreement}; Rater 3's 3 unranked items are excluded
from pairs involving Rater 3).

The averaged-LLM judge clears the inter-human ceiling ($\rho{=}+0.832$
vs.\ $+0.771$), so the LLM judges are noisy but unbiased estimators
of the same construct as the humans. Gemini-2.5-pro aligns marginally
better with humans than gpt-4o ($+0.803$ vs.\ $+0.793$), so it is the
safer single-judge proxy, but the averaged judge is what we use for
the headline distributional analyses in
Section~\ref{sec:incoherence-signature}.

\subsection{The Incoherence Signature}
\label{sec:incoherence-signature}

We rank-correlate every per-response aggregated feature
(Section~\ref{sec:metrics}) against (a)~the averaged-LLM rank per
(prompt,$T$) on all 500 prompts, and (b)~the human-majority rank on
the 150-prompt subset. For each feature we report the
\emph{within-prompt} Spearman $\rho$ across the three temperatures,
averaged across all prompts with at least two distinct feature
values.

\begin{figure}[t]
\centering
\begin{tikzpicture}
\begin{groupplot}[
  group style={group size=4 by 3, horizontal sep=1.0cm, vertical sep=1.0cm},
  width=3.2cm, height=2.7cm,
  xtick={0.3, 0.8, 1.5}, xticklabels={0.3, 0.8, 1.5},
  xmin=0.15, xmax=1.65,
  tick label style={font=\tiny},
  title style={font=\scriptsize, yshift=-1ex},
  grid=major, grid style={gray!20, line width=0.2pt},
  every axis plot/.append style={thick, mark size=1.5pt,
    mark options={fill=white, line width=0.8pt}},
]
\nextgroupplot[title={$n_{90}(p)$\,p10}, ymode=log]
\addplot[color=cblue, mark=*] coordinates {(0.3, 1.0) (0.8, 1.0) (1.5, 74.1)};
\nextgroupplot[title={$n_{95}(q)$\,p10}, ymode=log]
\addplot[color=cblue, mark=*] coordinates {(0.3, 1) (0.8, 1) (1.5, 131)};
\nextgroupplot[title={$n_{90}(q){-}n_{90}(p)$\,max}]
\addplot[color=cblue, mark=*] coordinates {(0.3, 0) (0.8, 0) (1.5, 102)};
\nextgroupplot[title={entropy\_shift\,p90}]
\addplot[color=cblue, mark=*] coordinates {(0.3, -0.003) (0.8, -0.003) (1.5, 0.603)};

\nextgroupplot[title={postCumOnPre$_{90}$\,mean}]
\addplot[color=cblue, mark=*] coordinates {(0.3, 1.000) (0.8, 0.982) (1.5, 0.868)};
\nextgroupplot[title={preCumOnPost$_{95}$\,p90}]
\addplot[color=cblue, mark=*] coordinates {(0.3, 1.000) (0.8, 1.000) (1.5, 0.984)};
\nextgroupplot[title={KL$(p\|q)$\,p10}, ymode=log]
\addplot[color=cblue, mark=*] coordinates {(0.3, 0.005) (0.8, 0.0005) (1.5, 0.008)};
\nextgroupplot[title={JS$(p,q)$\,p10}, ymode=log]
\addplot[color=cblue, mark=*] coordinates {(0.3, 0.0002) (0.8, 0.0001) (1.5, 0.0019)};

\nextgroupplot[title={Self-perplexity}, ymode=log, xlabel={$T$}, xlabel style={font=\tiny}]
\addplot[color=cred, mark=square*] coordinates {(0.3, 1.19) (0.8, 1.92) (1.5, 875)};
\nextgroupplot[title={Predictive entropy}, ymode=log, xlabel={$T$}, xlabel style={font=\tiny}]
\addplot[color=cred, mark=square*] coordinates {(0.3, 0.175) (0.8, 0.619) (1.5, 4.87)};
\nextgroupplot[title={Top-1 margin}, xlabel={$T$}, xlabel style={font=\tiny}]
\addplot[color=cred, mark=square*] coordinates {(0.3, 0.878) (0.8, 0.688) (1.5, 0.050)};
\nextgroupplot[title={Compression ratio}, xlabel={$T$}, xlabel style={font=\tiny}]
\addplot[color=cred, mark=square*] coordinates {(0.3, 0.515) (0.8, 0.527) (1.5, 0.675)};
\end{groupplot}

\node[color=cblue, font=\scriptsize\bfseries, rotate=90, anchor=center]
  at ($(group c1r1.west)!0.5!(group c1r2.west) + (-0.8cm, 0)$)
  {Distributional features};
\node[color=cred, font=\scriptsize\bfseries, rotate=90, anchor=center]
  at ($(group c1r3.west) + (-0.8cm, 0)$)
  {Literature baselines};
\end{tikzpicture}
\caption{Per-temperature mean values for the eight distributional
features (top two rows, blue) and four literature baselines (bottom
row, red). Every distributional feature step-changes at $T{=}1.5$
by one to three orders of magnitude. The mid-vs-bland call
($T{=}0.8$ vs $T{=}0.3$) is the closer one: the baselines show some
shift between the two coherent temperatures, but not enough to
translate into reliable within-prompt ranking
(Figure~\ref{fig:methods}). Log $y$-axis where values span multiple
orders of magnitude. Aggregator suffixes on panel titles:
\texttt{p10}/\texttt{p90}\,=\,10th/90th percentile across a trace;
\texttt{max}\,=\,maximum across a trace.}
\label{fig:metrics}
\end{figure}
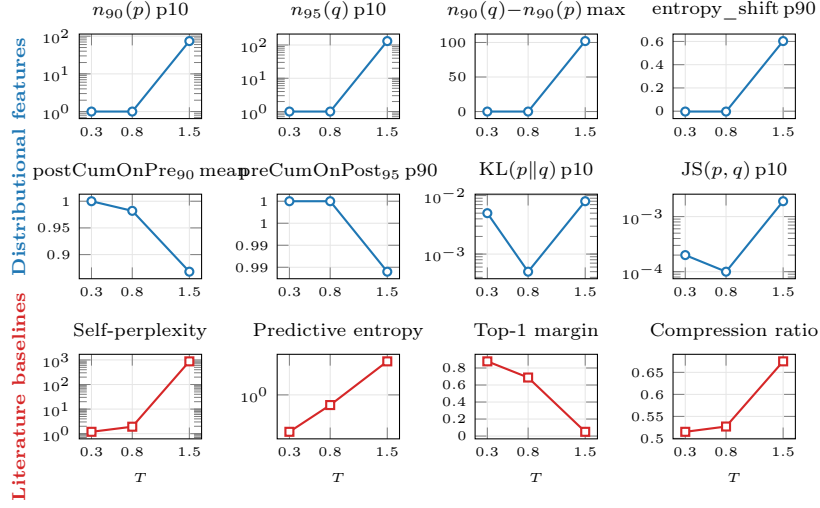

Figure~\ref{fig:metrics} brings together the per-$T$ mean values
for every metric we examine in this section: our distributional
features in the top two rows, and the four literature baselines we
compare against in Section~\ref{sec:vs-baselines} in the bottom
row. Every distributional feature cleanly separates $T{=}1.5$ from
the other two temperatures, often by one to three orders of
magnitude, and Table~\ref{tab:top-features} translates that
separation into Spearman correlations against the averaged-LLM
ground truth.

Table~\ref{tab:top-features} ranks the same distributional features
by how strongly they correlate with the averaged-LLM ground truth,
alongside their correlation against the human-majority ranking on
the 150-prompt subset. Within-prompt $\rho$ measures how reliably a
single feature value predicts the three-temperature ranking on a
given prompt.

\begin{table}[t]
\centering
\caption{Per-token distributional features ranked by within-prompt
Spearman $\rho$ against the averaged-LLM creativity rank
($n{=}500$ prompts). The ``human-maj $\rho$'' column reports the
same statistic against the three-rater human-majority ranking on
the $150$-prompt subset. Per-$T$ mean values are shown in
Figure~\ref{fig:metrics}. Aggregator suffixes: \texttt{\_\_p10}
(10th-percentile across the trace's tokens, i.e.\ the ``calmest''
position), \texttt{\_\_p90} (90th-percentile, ``most-perturbed''),
\texttt{\_\_max} (maximum across a trace).}
\label{tab:top-features}
\begin{tabular}{lccc}
\toprule
Feature & avg-LLM $\rho$ & human-maj $\rho$ & frac$\geq 0.6$ \\
\midrule
$n_{90}(p)$\texttt{\_\_p10}             & $\mathbf{+0.918}$ & $\mathbf{+0.870}$ & $1.00$ \\
$n_{90}(q)-n_{90}(p)$\texttt{\_\_max}   & $+0.917$ & $+0.869$ & $1.00$ \\
$n_{95}(q)$\texttt{\_\_p10}             & $+0.917$ & $+0.869$ & $1.00$ \\
entropy\_shift\texttt{\_\_p90}          & $+0.821$ & $+0.700$ & $0.74$ \\
$\mathrm{KL}(p\|q)$\texttt{\_\_p10}     & $+0.778$ & $+0.698$ & $0.72$ \\
$\mathrm{JS}$\texttt{\_\_p10}           & $+0.767$ & $+0.798$ & $0.66$ \\
\textsc{preCumOnPost95}\texttt{\_\_p90} & $-0.765$ & $-0.791$ & $0.64$ \\
\bottomrule
\end{tabular}
\end{table}

\paragraph{Mass leakage off the plausible set.}
$\textsc{postCumOnPre}_{90}$ is the fraction of post-temperature
mass that stays on the pre-temperature top-$90\%$ set
(Figure~\ref{fig:metrics}, \textsc{postCumOnPre}$_{90}$\,mean
panel). It sits at $1.000$ at $T{=}0.3$ and $0.982$ at $T{=}0.8$,
then drops to $0.868$ at $T{=}1.5$. The drop is not gradual:
$T{=}0.3$ and $T{=}0.8$ are nearly indistinguishable, then $T{=}1.5$
is a step away. Temperature at $T{=}1.5$ leaks about $13$\,pp of the
mass off the model's own pre-temperature plausible candidates onto
tokens outside that set.

\subsection{Comparison Against Four Literature Baselines}
\label{sec:vs-baselines}

\begin{figure}[t]
\centering
\begin{minipage}[t]{0.49\textwidth}
\centering
\begin{tikzpicture}[x=7cm, y=-0.5cm, font=\footnotesize]
\def\xmin{0.65}
\def\xmax{1.00}

\node[cblue, font=\scriptsize\bfseries, anchor=south]
  at ({(\xmin+\xmax)/2}, -1.0) {This work (distributional features)};

\foreach \xv in {0.65, 0.70, 0.75, 0.80, 0.85, 0.90, 0.95, 1.00} {%
  \draw[gray!25, line width=0.3pt] (\xv, 0) -- (\xv, 8.5);%
}

\foreach \row/\lbl in {%
    0/{$n_{90}(p)$ p10},%
    1/{$n_{90}(q){-}n_{90}(p)$ max},%
    2/{$n_{95}(q)$ p10},%
    3/{entropy\_shift p90}%
} {%
  \pgfmathsetmacro{\ymid}{\row*2 + 1}%
  \node[anchor=east, font=\scriptsize] at (\xmin-0.003, \ymid) {\lbl};%
}

\foreach \rho/\row in {0.918/0, 0.917/1, 0.917/2, 0.821/3} {%
  \pgfmathsetmacro{\yhi}{\row*2 + 0.3}%
  \pgfmathsetmacro{\ylo}{\row*2 + 0.95}%
  \fill[fill=cblue] (\xmin, \yhi) rectangle (\rho, \ylo);%
  \node[anchor=west, font=\tiny]
    at (\rho+0.004, {(\yhi+\ylo)/2})
    {\pgfmathprintnumber[fixed,fixed zerofill,precision=3]{\rho}};%
}

\foreach \rho/\row in {0.870/0, 0.869/1, 0.869/2, 0.700/3} {%
  \pgfmathsetmacro{\yhi}{\row*2 + 1.15}%
  \pgfmathsetmacro{\ylo}{\row*2 + 1.8}%
  \fill[fill=corange] (\xmin, \yhi) rectangle (\rho, \ylo);%
  \node[anchor=west, font=\tiny]
    at (\rho+0.004, {(\yhi+\ylo)/2})
    {\pgfmathprintnumber[fixed,fixed zerofill,precision=3]{\rho}};%
}

\draw[->, gray!70] (\xmin, 8.5) -- (\xmax+0.003, 8.5);
\foreach \xv in {0.70, 0.80, 0.90, 1.00} {%
  \draw[gray!70, line width=0.4pt] (\xv, 8.5) -- (\xv, 8.85);%
  \node[font=\scriptsize, anchor=north] at (\xv, 8.95) {\xv};%
}
\foreach \xv in {0.65, 0.75, 0.85, 0.95} {%
  \draw[gray!70, line width=0.3pt] (\xv, 8.5) -- (\xv, 8.7);%
}
\node[font=\footnotesize, anchor=north]
  at ({(\xmin+\xmax)/2}, 10.1) {Within-prompt Spearman $\rho$};

\begin{scope}[shift={(\xmin+0.02, 11.6)}]
  \fill[cblue]   (0, 0)   rectangle (0.012, 0.55);
  \node[anchor=west, font=\tiny] at (0.016, 0.275) {vs.\ averaged-LLM};
  \fill[corange] (0, 0.95) rectangle (0.012, 1.5);
  \node[anchor=west, font=\tiny] at (0.016, 1.225) {vs.\ human-majority};
\end{scope}
\end{tikzpicture}
\end{minipage}%
\hfill
\begin{minipage}[t]{0.49\textwidth}
\centering
\begin{tikzpicture}[x=7cm, y=-0.5cm, font=\footnotesize]
\def\xmin{0.65}
\def\xmax{1.00}

\node[cgray, font=\scriptsize\bfseries, anchor=south]
  at ({(\xmin+\xmax)/2}, -1.0) {Baselines};

\foreach \xv in {0.65, 0.70, 0.75, 0.80, 0.85, 0.90, 0.95, 1.00} {%
  \draw[gray!25, line width=0.3pt] (\xv, 0) -- (\xv, 8.5);%
}

\foreach \row/\lbl in {%
    0/{Self-perplexity},%
    1/{Predictive entropy},%
    2/{Top-1 margin ($|\rho|$)},%
    3/{Compression ratio}%
} {%
  \pgfmathsetmacro{\ymid}{\row*2 + 1}%
  \node[anchor=east, font=\scriptsize] at (\xmin-0.003, \ymid) {\lbl};%
}

\foreach \rho/\row in {0.753/0, 0.753/1, 0.753/2, 0.750/3} {%
  \pgfmathsetmacro{\yhi}{\row*2 + 0.3}%
  \pgfmathsetmacro{\ylo}{\row*2 + 0.95}%
  \fill[fill=cgray] (\xmin, \yhi) rectangle (\rho, \ylo);%
  \node[anchor=west, font=\tiny]
    at (\rho+0.004, {(\yhi+\ylo)/2})
    {\pgfmathprintnumber[fixed,fixed zerofill,precision=3]{\rho}};%
}

\foreach \rho/\row in {0.760/0, 0.760/1, 0.760/2, 0.734/3} {%
  \pgfmathsetmacro{\yhi}{\row*2 + 1.15}%
  \pgfmathsetmacro{\ylo}{\row*2 + 1.8}%
  \fill[fill=cgraylight] (\xmin, \yhi) rectangle (\rho, \ylo);%
  \node[anchor=west, font=\tiny]
    at (\rho+0.004, {(\yhi+\ylo)/2})
    {\pgfmathprintnumber[fixed,fixed zerofill,precision=3]{\rho}};%
}

\draw[->, gray!70] (\xmin, 8.5) -- (\xmax+0.003, 8.5);
\foreach \xv in {0.70, 0.80, 0.90, 1.00} {%
  \draw[gray!70, line width=0.4pt] (\xv, 8.5) -- (\xv, 8.85);%
  \node[font=\scriptsize, anchor=north] at (\xv, 8.95) {\xv};%
}
\foreach \xv in {0.65, 0.75, 0.85, 0.95} {%
  \draw[gray!70, line width=0.3pt] (\xv, 8.5) -- (\xv, 8.7);%
}
\node[font=\footnotesize, anchor=north]
  at ({(\xmin+\xmax)/2}, 10.1) {Within-prompt Spearman $\rho$};

\begin{scope}[shift={(\xmin+0.02, 11.6)}]
  \fill[cgray]      (0, 0)   rectangle (0.012, 0.55);
  \node[anchor=west, font=\tiny] at (0.016, 0.275) {vs.\ averaged-LLM};
  \fill[cgraylight] (0, 0.95) rectangle (0.012, 1.5);
  \node[anchor=west, font=\tiny] at (0.016, 1.225) {vs.\ human-majority};
\end{scope}
\end{tikzpicture}
\end{minipage}
\caption{Within-prompt Spearman $\rho$ for the top four distributional features
(left, blue/orange) and four literature baselines (right, gray) against the two
ground truths. The averaged-LLM column uses all 500 prompts; human-majority
uses the 150-prompt subset. Top-1 margin's negative sign is absorbed by
reporting $|\rho|$. Our top feature beats the best baseline by $+0.165$
(averaged-LLM) and $+0.110$ (human-majority).}
\label{fig:methods}
\end{figure}
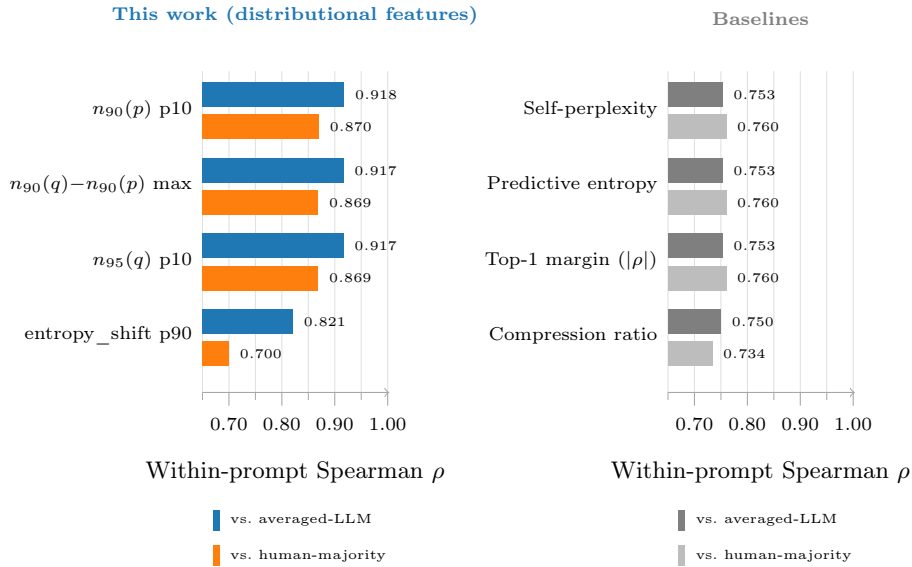

We compute four canonical reference-free baselines on the same 1500
traces and the same renormalised top-$K$ head:
\textbf{self-perplexity} \cite{jelinek1977perplexity};
mean post-temperature \textbf{predictive entropy} of the
chosen-token distribution \cite{malinin2021uncertainty}; mean
\textbf{top-1 margin} $q_{(1)}{-}q_{(2)}$; and \textbf{compression
ratio} $|\mathrm{gzip}(\text{text})|/|\text{text}|$
\cite{shaib2024standardizing,padmakumar2024writing}, the only
sequence-level baseline. Per-$T$ means for all four are in the
bottom row of Figure~\ref{fig:metrics}; their rank correlations
against each ground truth, alongside our distributional features,
are in Figure~\ref{fig:methods}.

All four baselines cluster tightly: $\Delta\rho{<}0.01$ on the
averaged-LLM panel and $\Delta\rho{=}0.026$ on human-majority
(Figure~\ref{fig:methods}, right panel). Our top distributional
feature beats the best baseline by $+0.165$ ($0.918$ vs.\ $0.753$)
against the averaged-LLM judge and by $+0.110$ ($0.870$ vs.\ $0.760$)
against the human majority.

\subsection{Where the Per-Token View Saturates}
\label{sec:negative-result}

The headline correlations above are driven by the clean
$T{=}1.5$-vs-rest separation. The finer mid-vs-bland distinction
($T{=}0.8$ vs $T{=}0.3$) sits closer to the limit of what any
single per-token feature in our catalogue can resolve, and the
literature baselines do not resolve it either. For our top feature
$n_{90}(p)$\texttt{\_\_p10} the per-$T$ mean is identical at
$T{=}0.3$ and $T{=}0.8$ ($1.0$ at both); for self-perplexity it
shifts modestly ($1.19$ vs.\ $1.92$); meanwhile the gap from either
to $T{=}1.5$ is one to three orders of magnitude on every metric. The signal that
distinguishes the two coherent regimes appears to live at the
sequence level (which content the model commits to over the
trajectory) rather than in how individual next-token distributions
are shaped at any one step. We return to this in
Section~\ref{sec:discussion}.
\section{Discussion}
\label{sec:discussion}

\subsection{Why the Pre-vs-Post View Beats Single-Temperature Scalars}
The headline finding $\rho{=}0.918$ against the averaged LLM judge
and $\rho{=}0.870$ against the human majority, versus
$|\rho|\!\approx\!0.76$ for every literature baseline we tested, is
not a small effect. The four baselines cluster within $\Delta\rho{<}0.01$
of each other on averaged-LLM and within $\Delta\rho{=}0.026$ on
human-majority; our top feature beats the best of them by $+0.165$
and $+0.110$ respectively, i.e.\ a margin many times larger than
the inter-baseline spread on either panel. Self-perplexity, predictive entropy, top-1
margin, and gzip compression ratio each collapse the same generation
to a single scalar; they cannot see that the temperature transform
$z \mapsto z/T$ reshapes the distribution before sampling. The
pre-vs-post view does see this, from the same forward pass, with no
extra inference cost.

The win comes from a mechanism the baselines structurally cannot
access. Across our measures (KL, TV, Hellinger, top-1-prob shift,
$n_\alpha$ width, mass leakage), the $T{=}1.5$ generations sit far
from $T{=}0.3$ and $T{=}0.8$ in distributional feature space, and
the shift is uniform across the trace: even the 10th-percentile
position moves sharply, not just occasional outliers. The cleanest
quantification is the per-token mass leakage off the pre-temperature
top-$90\%$ plausible set, which drops from $0.982$ at $T{=}0.8$ to
$0.868$ at $T{=}1.5$ (a ${\sim}13$-percentage-point leak). The tail of
the model's belief carries enough mass at $T{=}1.5$ that the
sampled tokens are no longer recognisable continuations of the
prompt's context, qualitatively consistent with the typicality
framing of Meister et~al.\ \cite{meister2023typical} and the
nucleus-sampling intuition \cite{holtzman2020curious}.

\subsection{Scope of the Per-Token View}
The pre-vs-post view's strongest signal is at the incoherence
boundary. The mid-vs-bland gap ($T{=}0.8$ vs $T{=}0.3$), where both
LLM and human judges find the distinction subtle, sits closer to the
limit of what any single per-token statistic can resolve: at
$T{=}0.3$ and $T{=}0.8$, $p$ and $q$ preserve roughly the same shape
at every position, so the signal lives instead in \emph{which token
gets sampled} at flat-distribution positions, whose effect compounds
over the trajectory. This points toward sequence-level features
(n-gram diversity, content novelty, self-similarity across samples)
as a complementary layer to the per-token catalogue, not as a
replacement for it.

\subsection{Implications for Reference-Free Metric Design}
A creativity-aware, reference-free metric was proposed in an earlier
version of this work \cite{parupudi2025confidence} using a
hand-designed scalar (chosen-token probability times the top-$N$
standard deviation). The results here suggest that the ``one scalar
per response'' framing may be too crude. Information about the
temperature regime is layered: which features change, by how much,
and at which positions of the trace. Different aggregators carry
the signal for different feature families: \texttt{p10} (the calmest
position) for the cumulative-mass-width levels, \texttt{max} (the
single most-perturbed position) for the flattening difference, and
\texttt{p90} (the most-perturbed position) for the signed entropy
shift and \textsc{preCumOnPost} (Table~\ref{tab:top-features}). A
single scalar collapses that structure away.

\section{Limitations}
\label{sec:limitations}

\paragraph{Single model.}
All distributional analysis is on Llama-3.1-8B-Instruct \cite{grattafiori2024llama3}. On the mechanistic argument (temperature transform + tail-mass redistribution is architecture-agnostic) we expect qualitative replication on other architectures.

\paragraph{Top-$K$ truncation.}
All divergences are computed on the renormalised top-$K$ head, $p$ and $q$ are both probability distributions over the same $K$ captured tokens. This is exact (no estimation) and consistent across all three temperatures, so comparisons within-prompt and within-$T$ are apples-to-apples. However, we do not report absolute divergence magnitudes against any external benchmark.

\paragraph{Single-pass measurement.}
We measure $p$ and $q$ from a single forward pass per generated token; we do not, for example, run the model multiple times to characterise sampling variance in the distribution itself. This is a conscious choice (single-pass is cheap and matches deployment), but it means we can't decompose the residual variance into "the model is uncertain about $p$ here" vs "$p$ is sharp but the trace diverges from baseline anyway".
\section{Conclusion}
\label{sec:conclusion}

We studied reference-free evaluation of open-ended LLM generation
by treating the distributional shift caused by sampling temperature
as the primary observable. This let us ask not ``how confident was
the model?'' but ``how far did temperature push the sampling
distribution away from the model's own belief, and where?''.

\begin{enumerate}
  \item \textbf{A single per-token feature derived from the
        pre-vs-post temperature reshape beats every standard
        reference-free baseline on both ground truths.} The top
        feature predicts the within-prompt creativity rank at
        Spearman $\rho{=}0.918$ against an averaged LLM judge and
        $\rho{=}0.870$ against the human majority; the best of
        self-perplexity, predictive entropy, top-1 margin, and gzip
        compression ratio tops out at $|\rho|\!\approx\!0.76$ on
        both panels. The pre-vs-post divergence catalogue is a
        cheap, qualitatively interpretable single-pass measurement
        that requires no reference text and avoids the multi-sample
        cost of methods like semantic entropy \cite{kuhn2023semantic}.
  \item \textbf{The mechanism is a sharp distributional signature of
        the incoherence regime.} At $T{=}1.5$, mass on the
        pre-temperature top-$90\%$ plausible set drops from $0.982$
        at $T{=}0.8$ to $0.868$ (a ${\sim}13$-percentage-point leak),
        and $n_{95}(q)$ inflates by an order of magnitude. The
        distortion is uniform across the trace, not confined to
        occasional ``creative'' tokens.
  \item \textbf{Per-token aggregates do not separate $T{=}0.8$ from
        $T{=}0.3$.} The mid-vs-bland distinction lives at the
        sequence level (which token gets sampled at
        flat-distribution positions, compounded over the trajectory)
        and is a natural target for sequence-level features rather
        than a limit of the pre-vs-post view itself.
\end{enumerate}

\paragraph{Future directions.}
Two immediate extensions: (a)~test the pre-vs-post catalogue on
other model families (Mistral, Qwen, larger Llama) to check whether
the incoherence signature is architecture-agnostic, as the
mechanistic argument suggests; and (b)~run a finer temperature
sweep on a smaller prompt subset to characterise the
$T{=}0.8 \to T{=}1.5$ transition more precisely.

{\footnotesize
\bibliographystyle{splncs04}
\bibliography{refs}

\begin{thebibliography}{10}
\providecommand{\url}[1]{\texttt{#1}}
\providecommand{\urlprefix}{URL }
\providecommand{\doi}[1]{https://doi.org/#1}

\bibitem{achiam2023gpt}
Achiam, J., et~al.: {GPT-4} technical report (2023), arXiv:2303.08774

\bibitem{brown1993statistical}
Brown, P.F., Pietra, V.J.D., Pietra, S.A.D., Mercer, R.L.: The mathematics of statistical machine translation: parameter estimation. Comput. Linguist.  \textbf{19}(2),  263–311 (Jun 1993)

\bibitem{chakrabarty2024art}
Chakrabarty, T., Laban, P., Agarwal, D., Muresan, S., Wu, C.S.: Art or artifice? {L}arge language models and the false promise of creativity. In: Proceedings of the 2024 {CHI} Conference on Human Factors in Computing Systems (2024), arXiv:2309.14556

\bibitem{fabbri2021summeval}
Fabbri, A.R., Kry{\'s}ci{\'n}ski, W., McCann, B., Xiong, C., Socher, R., Radev, D.: {SummEval}: Re-evaluating summarization evaluation. Transactions of the Association for Computational Linguistics  \textbf{9},  391--409 (2021)

\bibitem{fan2018hierarchical}
Fan, A., Lewis, M., Dauphin, Y.: Hierarchical neural story generation. In: Proceedings of the 56th Annual Meeting of the Association for Computational Linguistics (ACL). pp. 889--898 (2018), arXiv:1805.04833

\bibitem{farquhar2024detecting}
Farquhar, S., Kossen, J., Kuhn, L., Gal, Y.: Detecting hallucinations in large language models using semantic entropy. Nature  \textbf{630},  625--630 (2024), \url{https://doi.org/10.1038/s41586-024-07421-0}

\bibitem{fu2024gptscore}
Fu, J., Ng, S.K., Jiang, Z., Liu, P.: {GPTS}core: Evaluate as you desire. In: Duh, K., Gomez, H., Bethard, S. (eds.) Proceedings of the 2024 Conference of the North American Chapter of the Association for Computational Linguistics: Human Language Technologies (Volume 1: Long Papers). pp. 6556--6576. Association for Computational Linguistics, Mexico City, Mexico (Jun 2024). \doi{10.18653/v1/2024.naacl-long.365}, \url{https://aclanthology.org/2024.naacl-long.365/}

\bibitem{grattafiori2024llama3}
Grattafiori, A., et~al.: The llama 3 herd of models (2024), arXiv:2407.21783; meta-llama/Llama-3.1-8B-Instruct

\bibitem{guilford1967nature}
Guilford, J.P.: The Nature of Human Intelligence. McGraw-Hill (1967)

\bibitem{holtzman2020curious}
Holtzman, A., Buys, J., Du, L., Forbes, M., Choi, Y.: The curious case of neural text degeneration. In: International Conference on Learning Representations (ICLR) (2020), arXiv:1904.09751

\bibitem{jelinek1977perplexity}
Jelinek, F., Mercer, R.L., Bahl, L.R., Baker, J.M.: Perplexity—a measure of the difficulty of speech recognition tasks. Journal of the Acoustical Society of America  \textbf{62} (1977), \url{https://api.semanticscholar.org/CorpusID:121680873}

\bibitem{kim2024prometheus}
Kim, S., Shin, J., Cho, Y., Jang, J., Longpre, S., Lee, H., Yun, S., Shin, S., Kim, S., Thorne, J., Seo, M.: {Prometheus}: Inducing fine-grained evaluation capability in language models. In: International Conference on Learning Representations (ICLR) (2024), arXiv:2310.08491

\bibitem{kuhn2023semantic}
Kuhn, L., Gal, Y., Farquhar, S.: Semantic uncertainty: Linguistic invariances for uncertainty estimation in natural language generation. In: International Conference on Learning Representations (ICLR) (2023), arXiv:2302.09664

\bibitem{li2016diversity}
Li, J., Galley, M., Brockett, C., Gao, J., Dolan, B.: A diversity-promoting objective function for neural conversation models. In: Proceedings of the 2016 Conference of the North American Chapter of the Association for Computational Linguistics (NAACL-HLT). pp. 110--119 (2016), arXiv:1510.03055

\bibitem{liu2023geval}
Liu, Y., Iter, D., Xu, Y., Wang, S., Xu, R., Zhu, C.: {G-Eval}: {NLG} evaluation using {GPT-4} with better human alignment. In: Proceedings of the 2023 Conference on Empirical Methods in Natural Language Processing (EMNLP) (2023), arXiv:2303.16634

\bibitem{malinin2021uncertainty}
Malinin, A., Gales, M.: Uncertainty estimation in autoregressive structured prediction. In: International Conference on Learning Representations (ICLR) (2021), arXiv:2002.07650

\bibitem{manakul2023selfcheckgpt}
Manakul, P., Liusie, A., Gales, M.J.F.: {SelfCheckGPT}: Zero-resource black-box hallucination detection for generative large language models. In: Proceedings of the 2023 Conference on Empirical Methods in Natural Language Processing (EMNLP) (2023), arXiv:2303.08896

\bibitem{meister2023typical}
Meister, C., Pimentel, T., Wiher, G., Cotterell, R.: Locally typical sampling. Transactions of the Association for Computational Linguistics  \textbf{11},  102--121 (2023), arXiv:2202.00666

\bibitem{padmakumar2024writing}
Padmakumar, V., He, H.: Does writing with language models reduce content diversity? In: International Conference on Learning Representations (ICLR) (2024), arXiv:2309.05196

\bibitem{papineni2002bleu}
Papineni, K., Roukos, S., Ward, T., Zhu, W.J.: {BLEU}: a method for automatic evaluation of machine translation. In: Proceedings of the 40th Annual Meeting of the Association for Computational Linguistics (ACL). pp. 311--318 (2002)

\bibitem{parupudi2025confidence}
Parupudi, R.: Confidence, not perplexity: A better metric for the creative era of {LLMs} (2025), arXiv:2510.08596

\bibitem{shaib2024standardizing}
Shaib, C., Barrow, J., Sun, J., Siu, A.F., Wallace, B.C., Nenkova, A.: Standardizing the measurement of text diversity: A tool and a comparative analysis of scores. arXiv preprint  (2024), arXiv:2403.00553

\bibitem{stevenson2022putting}
Stevenson, C., Smal, I., Baas, M., Grasman, R., van~der Maas, H.: Putting {GPT-3}'s creativity to the (alternative uses) test. In: Proceedings of the 13th International Conference on Computational Creativity (ICCC) (2022), arXiv:2206.08932

\bibitem{wang2023selfconsistency}
Wang, X., Wei, J., Schuurmans, D., Le, Q.V., Chi, E.H., Narang, S., Chowdhery, A., Zhou, D.: Self-consistency improves chain of thought reasoning in language models. In: International Conference on Learning Representations (ICLR) (2023), arXiv:2203.11171

\bibitem{yuan2021bartscore}
Yuan, W., Neubig, G., Liu, P.: {BARTScore}: Evaluating generated text as text generation. In: Advances in Neural Information Processing Systems (NeurIPS) (2021)

\bibitem{zellers2019hellaswag}
Zellers, R., Holtzman, A., Bisk, Y., Farhadi, A., Choi, Y.: {HellaSwag}: Can a machine really finish your sentence? In: Proceedings of the 57th Annual Meeting of the Association for Computational Linguistics (ACL) (2019), arXiv:1905.07830

\bibitem{zhang2020bertscore}
Zhang, T., Kishore, V., Wu, F., Weinberger, K.Q., Artzi, Y.: {BERTScore}: Evaluating text generation with {BERT}. In: International Conference on Learning Representations (ICLR) (2020), arXiv:1904.09675

\bibitem{zhu2018texygen}
Zhu, Y., Lu, S., Zheng, L., Guo, J., Zhang, W., Wang, J., Yu, Y.: Texygen: A benchmarking platform for text generation models. In: Proceedings of the 41st International ACM SIGIR Conference (2018), arXiv:1802.01886

\end{thebibliography}
}
\end{document}